\documentclass[11pt,a4paper]{article}
\usepackage[hyperref]{acl2020}
\usepackage{times}
\usepackage{latexsym}

\usepackage{graphicx}
\usepackage{amsmath}
\usepackage{amsfonts}
\usepackage{amssymb}
\usepackage{multirow}

\usepackage{microtype}

\aclfinalcopy 

\newcommand\IMSone{\texttt{IMS-CUB1}}
\newcommand\IMStwo{\texttt{IMS-CUB2}}

\title{The IMS--CUBoulder System for the SIGMORPHON 2020 Shared Task on Unsupervised Morphological Paradigm Completion}

\author{ Manuel Mager \\
  Institute for Natural Language Processing\\
  University of Stuttgart \\
  \texttt{manuel.mager@ims.uni-stuttgart.de} \\\And
  Katharina Kann \\
  University of Colorado Boulder \\
  \texttt{katharina.kann@colorado.edu} \\}

\date{}

\begin{document}
\maketitle
\begin{abstract}
In this paper, we present the systems of the University of Stuttgart IMS and the University of Colorado Boulder (IMS--CUBoulder) for SIGMORPHON 2020 Task 2 on unsupervised morphological paradigm completion \cite{kann-etal-2020-sigmorphon}. The task consists of generating the morphological paradigms of a set of lemmas, given only the lemmas themselves and unlabeled text. 
Our proposed system is a modified version of the baseline introduced together with the task.
In particular, we experiment with substituting the inflection generation component with an LSTM sequence-to-sequence model and an LSTM pointer-generator network. 
Our pointer-generator system obtains the best score of all seven submitted systems on average over all languages, and outperforms the official baseline, which was best overall, on Bulgarian and Kannada. 
\end{abstract}

\section{Introduction}
\begin{figure}[t]
    \centering
    \includegraphics[width=.85\columnwidth]{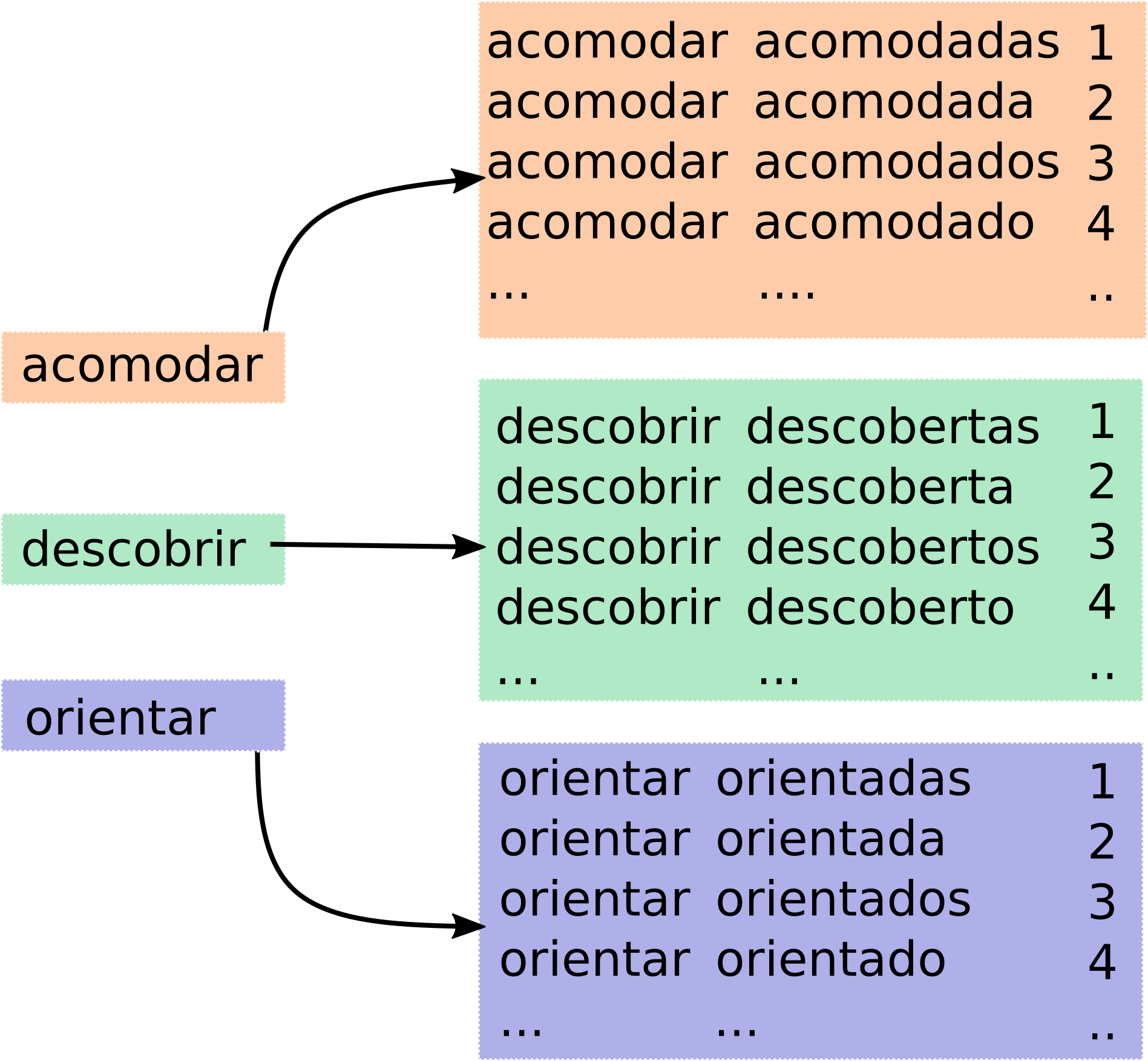}
    \caption{Partial Portuguese development examples. The input is a list of lemmas, and the output is a list of all inflected forms of each lemma. In this example, unnamed paradigm slots correspond to the following UniMorph features: 1=V.PTCP;FEM;PL;PST, 2=V.PTCP;FEM;SG;PST, 3=V.PTCP;MASC;PL;PST, 4=V.PTCP;MASC;SG;PST.}
    \label{fig:example}
\end{figure}{}
In recent years, a lot of progress has been made on the task of morphological inflection, which consists of generating an inflected word, given a lemma and a list of morphological features \cite{kann-schutze-2017-lmu,makarov2018imitation,cotterell-etal-2016-sigmorphon, cotterell-etal-2017-conll,cotterell-etal-2018-conll,mccarthy2019sigmorphon}. The systems developed for this task learn to model inflection in morphologically complex languages in a supervised fashion. 

However, not all languages have annotated data available. For the 2018 SIGMORPHON shared task \cite{cotterell-etal-2018-conll}, data for $103$ unique languages has been provided. Even this highly multilingual dataset is just covering $1.61\%$ of the $6359$ languages\footnote{The number of languages can vary depending on the classification schema used.} that exist in the world \cite{lewis2009ethnologue}. The unsupervised morphological paradigm completion task \cite{jin2020unsupervised} aims at generating inflections -- more specifically all inflected forms, i.e., the entire paradigms, of given lemmas -- without any explicit morphological information during training. A system that is able to solve this problem can generate morphological resources for most of the world's languages easily.  This motivates us to participate in the SIGMORPHON 2020 shared task on unsupervised morphological paradigm completion \cite{kann-etal-2020-sigmorphon}.

The task, however, is challenging: As the number of inflected forms per lemma is unknown a priori, an unsupervised morphological paradigm completion system needs to detect the paradigm size from raw text. Since the names of morphological features expressed in a language are not known if there is no supervision, a system should mark which inflections correspond to the same morphological features across lemmas, but needs to do so without using names, cf. Figure \ref{fig:example}. 
For the shared task, no 
external resources such as pretrained models, annotated data, or even additional monolingual text can be used. The same holds true for multilingual models.

We submit two systems, which are both modifications of the official shared task baseline. The latter is a pipeline system, which performs four steps: edit tree retrieval, additional lemma retrieval, paradigm size discovery, and inflection generation \cite{jin2020unsupervised}. We experiment with substituting the original generation component, which is either a simple non-neural system \cite{cotterell-etal-2017-conll} or a transducer-based hard-attention model \cite{makarov2018imitation} with an LSTM encoder-decoder architecture with attention \cite{bahdanau2015neural} -- \IMSone -- and a pointer-generator network \cite{see-etal-2017-get} -- \IMStwo. \IMStwo{}  achieves the best results of all submitted systems, outperforming the second best system by $2.07\%$ macro-averaged best-match accuracy \citep[BMAcc;][]{jin2020unsupervised}, when averaged over all languages. However, we underperform the baseline system, which performs $1.03\%$ BMAcc better than \IMStwo{}. Looking at individual languages, \IMStwo{} obtains the best results overall for Bulgarian and Kannada. 

The findings from our work on the shared task are as follows: i) the copy capabilities of a pointer-generator network are useful in this setup; and ii) unsupervised morphological paradigm completion is a challenging task:  no submitted system outperforms the baselines.

\section{Related Work}
Unsupervised methods have shown to be effective for morphological surface segmentation. LINGUISTICA \cite{Goldsmith:2001:ULM:972667.972668} and MORFESSOR \cite{creutz-2003-unsupervised,creutz2007unsupervised, poon2009unsupervised} are two unsupervised systems for the task.  

In the realm of morphological generation, \citet{yarowsky-wicentowski-2000-minimally} worked on a task which was similar to unsupervised morphological paradigm completion, but required additional  knowledge (e.g., a list of morphemes). \citet{dreyer-eisner-2011-discovering} used a set of seed paradigms to train a paradigm completion model. \citet{ahlberg-etal-2015-paradigm} and \citet{hulden-etal-2014-semi} also relied on information about the paradigms in the language. 
\citet{erdmann2020paradigm} proposed a system for a task similar to this shared task.

Learning to generate morphological paradigms in a fully supervised way is the more common approach. Methods include  \citet{durrett-denero-2013-supervised}, \citet{nicolai-etal-2015-inflection}, and \citet{kann-schutze-2018-neural}. Supervised morphological inflection has further gained popularity through previous SIGMORPHON and CoNLL--SIGMORPHON shared tasks on the topic \cite{cotterell-etal-2016-sigmorphon,cotterell-etal-2017-conll,cotterell-etal-2018-conll,mccarthy2019sigmorphon}. The systems proposed for these shared tasks have a special relevance for our work, as we investigate the performance of morphological inflection components based on \citet{kann2016med,kann-schutze-2016-single} and \citet{sharma2018iit}  within a pipeline for unsupervised morphological paradigm completion.

\begin{table}
    \centering
    \setlength{\tabcolsep}{3.5pt}
    \begin{tabular}{c|r r r}
Language & Training & Development & Test\\\hline 
Basque   & 85 & 16 & 499\\ 
Bulgarian& 1609 & 441 & 2874\\
English  & 343 & 83& 302\\ 
Finnish  & 2306 & 522& 1789\\ 
German   & 3940 & 999& 667\\ 
Kannada  & 832 & 211& 2854\\ 
Navajo   & 17 & 4 & 279\\ 
Spanish  & 1940 & 494 & 2506\\ 
Turkish  & 3095 & 787& 8502\\
    \end{tabular}
    \caption{Number of instances retrieved by steps 1 to 3 in our pipeline, which are used for training and development of our inflection generation components. The test set contains the lemma and paradigm slot for forms that need to be generated.}
    \label{tab:training_instances}
\end{table}{}

\section{System Description}\label{sec:baseline}
In this section, we introduce our pipeline system for unsupervised morphological paradigm completion. First, we describe the baseline system, since we rely on some of its components. Then, we describe our morphological inflection models. 

\begin{figure*}
    \centering
    \includegraphics[width=.95\linewidth]{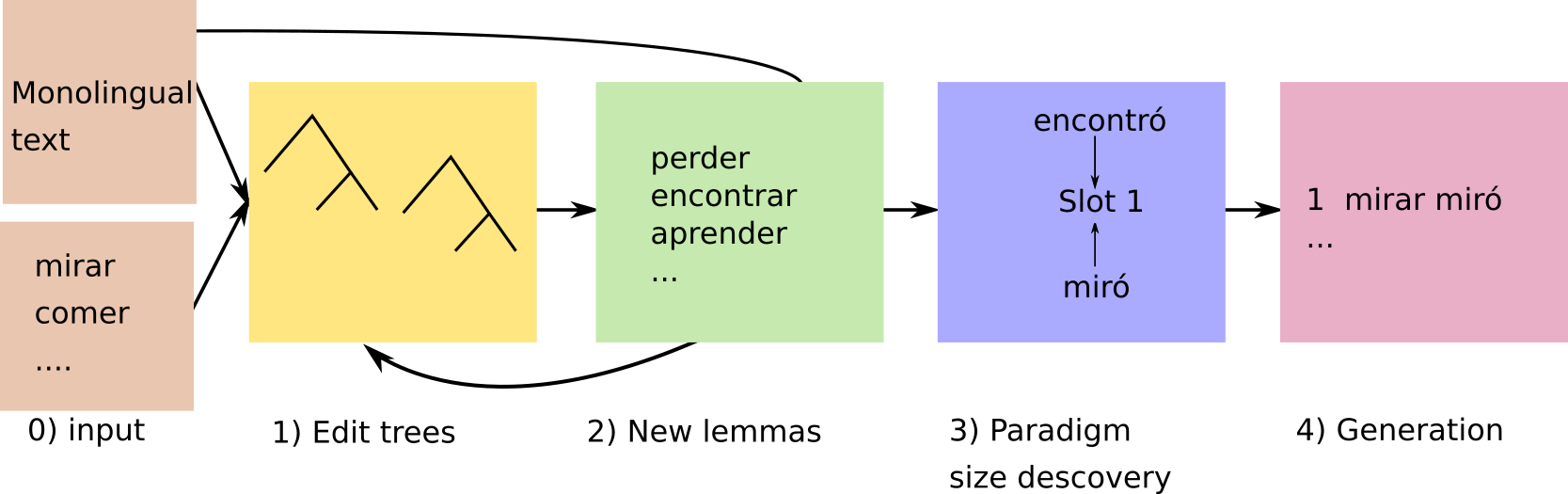}
    \caption{The baseline system. This paper experiments with modifying the generation module. All components are described in \S\ref{subsec:baseline}.}
    \label{fig:pipeline}
\end{figure*}
\subsection{The Shared Task Baseline}
\label{subsec:baseline}
For the initial steps of our pipeline, we employ the first three components of the baseline \citep{jin2020unsupervised}, cf. Figure \ref{fig:pipeline}, which we describe in this subsection. We use the official implementation.\footnote{\url{https://github.com/cai-lw/morpho-baseline}}

\paragraph{Retrieval of relevant edit trees.} This component (cf. Figure \ref{fig:pipeline}.1) identifies words in the monolingual corpus that could belong to a given lemma's paradigm by computing the longest common substring between the lemma and all words. Then, the transformation from a lemma to each word potentially from its paradigm is represented by edit trees \cite{chrupala2008towards}. Edit trees with frequencies are below a threshold are discarded.

\paragraph{Retrieval of additional lemmas.} To increase the confidence that retrieved edit trees represent valid inflections, more lemmas are needed (cf. Figure \ref{fig:pipeline}.2). To find those, the second component of the system applies edit trees to potential lemmas in the corpus. If enough potential inflected forms are found in the corpus, a lemma is considered valid. 

\paragraph{Paradigm size discovery.} Now the system needs to find a mapping between edit trees and paradigms (cf. Figure \ref{fig:pipeline}.3). This is done based on two assumptions: that for each lemma a maximum of one edit tree per paradigm slot can be found, and that each edit tree only realizes one paradigm slot for all lemmas. In addition, the similarity of potential slots is measured.
With these elements, similar potential slots are merged until the final paradigm size for a language is being determined.

\paragraph{Generation.} Now, that the system has a set of lemmas and corresponding potential inflected forms, the baseline employs a morphological inflection component, which learns to generate inflections from lemmas and a slot indicator, and generates missing forms (cf. Figure \ref{fig:pipeline}.4).
We experiment with substituting this final component.

In the remainder of this paper, we will refer to the original baselines with the non-neural system from \newcite{cotterell-etal-2017-conll} and the inflection model from \newcite{makarov2018imitation} 
as
\texttt{BL-1}  and \texttt{BL-2}, respectively.

\subsection{LSTM Encoder-Decoder}

We use an LSTM encoder-decoder model with attention \cite{bahdanau2015neural} for our first system, \IMSone, since it has been shown to obtain high performance on morphological inflection  \cite{kann2016med}. This model takes two inputs: a sequence of characters 
and a sequence of morphological features. It then generates the sequence of characters of the inflected form. 
For the input, we simply concatenate the paradigm slot number and all characters. 

\subsection{Pointer-Generator Network}
For \IMStwo, we use a pointer-generator network \cite{see-etal-2017-get}.\footnote{We use the following implementation: \url{https://github.com/abhishek0318/conll-sigmorphon-2018}} We expect this system to perform better than \IMSone{}, given the pointer-generator's better performance on morphological inflection in the low-resource setting \cite{sharma2018iit}. 
A pointer-generator network is a hybrid between an attention-based sequence-to-sequence model \cite{bahdanau2015neural} and a pointer network \cite{vinyals2015pointer}.

The standard pointer-generator network consists of a bidirectional LSTM \cite{hochreiter1997long} encoder and a unidirectional LSTM decoder with a copy mechanism.
Here, we follow \cite{sharma2018iit} and use two separate encoders: one for the lemma and one for the morphological tags. The decoder then computes the probability distribution of the output at each time step as a weighted sum of the probability distribution over the output vocabulary and the attention distribution over the input characters. The weights can be seen as the probability to generate or copy, respectively, and are computed by a feedforward network, given the last decoder hidden state. For details, we refer the reader to \newcite{sharma2018iit}.

\section{Experimental Setup}
\subsection{Data and Languages}
The shared task organizers provide data for five development languages, for which development sets with gold solutions are given. Those languages -- Maltese, Persian, Portuguese, Russian, Swedish -- are not taken into account for the final evaluation.

The test languages, in contrast, are supposed to be only for system evaluation and do not come with developments sets. For those languages -- Basque, Bulgarian, English, Finnish, German, Kannada, Navajo, Spanish, and Turkish -- only a list of lemmas and a monolingual Bible \cite{mccarthy-EtAl:2020:LREC1} are given.
\label{sec:exp_setup}
\begin{table}
    \centering
\setlength{\tabcolsep}{7.0pt}
    \begin{tabular}{c|r r r}
    &\multicolumn{3}{c}{\texttt{IMS-CUB}} \\
Language   & \texttt{1} & {\texttt{2-S}} & \texttt{2-V}\\\hline 
Basque     &  \bf25.00 & 18.75    &12.50 \\
Bulgarian  & 97.73     & \bf98.19 &97.28 \\
English    & 96.39     & \bf98.80 &\bf98.80 \\
Finnish    & \bf99.04  & 98.47    &98.85 \\
German     & 91.49     & \bf93.39 &91.99 \\
Kannada    & 91.47     & \bf92.89 &91.00 \\
Maltese    & 79.17     & 79.17    &\bf85.42 \\
Navajo     & 0.00      & 75.00    &\bf100.00 \\
Persian    & 95.56     & 94.81    &\bf95.56 \\
Portuguese & 93.81     & \bf93.87 &93.74 \\
Russian    & 92.15     & 93.02    &\bf93.19 \\
Spanish    & 92.91     & 92.71    &\bf93.52 \\
Swedish    & 93.48     & \bf93.69 &93.27 \\
Turkish    & 93.90     & 95.30    &\bf95.68 \\
    \end{tabular}
    \caption{Accuracy of our morphological inflection components on the development sets produced by the first three steps in our pipeline. We list both development and test languages.}
    \label{tab:generation_results}
\end{table}{}
\subsection{Evaluation Metric}
The official evaluation metric of the shared task is BMAcc \cite{jin2020unsupervised}. Gold solutions are obtained from UniMorph \cite{kirov2018unimorph}. Two versions of BMAcc exist: micro-averaged BMAcc and macro-averaged BMAcc. In this paper, we only report macro-averaged BMAcc, the  official shared task metric. 

During the development of our morphological generation systems, we use regular accuracy, the standard evaluation metric for morphological inflection \cite{cotterell-etal-2016-sigmorphon}.

\begin{table*}[h]
    \centering
\setlength{\tabcolsep}{7.0pt}
    \begin{tabular}{l | r r | r r | r r | r r r}
    & \multicolumn{2}{c |}{\texttt{BL}} & \multicolumn{2}{c|}{\texttt{KU-CST}} & \multicolumn{2}{c|}{\texttt{IMS-CUB}} & \multicolumn{3}{c}{\texttt{NYU-CUB}} \\
Language & \texttt{1}&	\texttt{2}&	\texttt{1}&\texttt{2}	& \texttt{1} & \texttt{2} & \texttt{1}&	\texttt{2}&	\texttt{3}	\\\hline
Basque   & 0.06&	     0.06&	0.02&	 0.01&	 0.04&	    00.06&	 0.05&	 0.05&	\bf  \underline{0.07}\\
Bulgarian&28.30&	    31.69&	2.99&	 4.15&	27.22&	\bf \underline{32.11}&	27.69&	28.94&	    27.89\\
English  &65.60&	\bf 66.20&	3.53&	17.29&	47.80&	    \underline{61.00}&	50.20&	52.80&	    51.20\\
Finnish  &05.33&	\bf  5.50&	0.39&	 2.08&	04.90&	    \underline{05.38}&	 5.36&	 5.47&	    05.35\\
German   &28.35&	\bf 29.00&	0.70&	 4.98&	24.60&	    \underline{28.35}&	27.30&	27.35&	    27.35\\
Kannada  &15.49&	    15.12&	4.27&	 1.69&	10.50&	\bf \underline{15.65}&	11.10&	11.16&	    11.10\\
Navajo   & 3.23&	\bf  3.27&	0.13&	 0.20&	 0.33&	    \underline{01.17}&	 0.40&	 0.43&	     0.43\\
Spanish  &22.96&	\bf 23.67&	3.52&	10.84&	19.50&	    \underline{22.34}&	20.39&	20.56&	    20.30\\
Turkish  &14.21&	\bf 15.53&	0.11&	 0.71&	13.54&	    14.73&	14.88&	\underline{15.39}&	    15.13\\\hline
Average  &20.39&	\bf 21.12&	1.74&	04.66&	16.49&	    \underline{20.09}&	17.49&	18.02&	    17.65\\
    \end{tabular}
    \caption{Final performance (macro-average BMAcc in percentages) of all systems on all test languages. Best scores overall are in bold, and best scores of submitted systems are underlined.}
    \label{tab:results}
\end{table*}

\subsection{Morphological Inflection Component}
\paragraph{Morphological inflection data.}
We use the first three components of the baseline model, i.e., the ones performing edit tree retrieval, additional lemma retrieval, and paradigm size discovery, to create training and development data for our inflection models. Those datasets consist of lemma--inflection pairs found in the raw text, together with a number indicating the (predicted) paradigm slot, and are described in Table \ref{tab:training_instances}. 

The test set for our morphological inflection systems consist of the lemma--paradigm slot pairs not found in the corpus.

\paragraph{Hyperparameters.}
For \IMSone{}, we use an embedding size of 300, a hidden layer of size 100, a batch size of 20, Adadelta \cite{zeiler2012adadelta} for optimization, and a learning rate of 1. For each language, we train a system for 100 epochs, using early stopping with a patience of 10 epochs.

For \IMStwo{}, we follow two different approaches. The first is to use a single hyperparameter configuration for all languages (\IMStwo-\texttt{S}). The second consists of using a variable setup depending on the training set size (\IMStwo-\texttt{V}). For \IMStwo-\texttt{S}, we use an embedding size of 300, a hidden layer size of 100, a dropout rate of 0.3, and train for 60 epochs with an early-stopping patience of 10 epochs. We further use an Adam \cite{kingma2014adam} optimizer with an initial learning rate of 0.001. 

For \IMStwo-\texttt{V}, we use the following hyperparameters for training set size $T$:
\begin{itemize}
    \item \textbf{$T < 101$}: an embedding size of 100, a dropout coefficient of 0.5, 300 epochs of training, and an early-stopping patience of 100;
    \item \textbf{$100 < T < 501$}: an embedding size of 100, a dropout coefficient of 0.5, 80 training epochs, and an early-stopping patience of 20;
    \item \textbf{$500 < T$ }: the same hyperparameters as for \IMStwo-\texttt{S}. 
\end{itemize}{}
For \IMStwo{}, we select the best performing system (between \IMStwo-\texttt{S} and \IMStwo-\texttt{V}) as our final model. The models are evaluated on the morphological inflection task development set using accuracy. All scores are shown in Table \ref{tab:generation_results}.

\subsection{Results}
Table \ref{tab:results} shows the official test set results for \IMSone{} and \IMStwo{}, compared to the official baselines and all other submitted systems. 

Our best system, \IMStwo, achieves the highest scores of all submitted systems (i.e., excluding the baselines), outperforming the second best submission by $2.07\%$ BMAcc. However, \texttt{BL-1} and \texttt{BL-2} outperform  \IMStwo{} by $1.03\%$ and $0.3\%$, respectively.
Looking at the results for individual languages, \IMStwo{} obtains the highest performance overall for Bulgarian (difference to the second best system $0.42\%$) and Kannada (difference to the second best system $0.53\%$).  
Comparing our two submissions, \IMSone{} underperforms \IMStwo{} by $3.6\%$, showing that vanilla sequence-to-sequence models are not optimally suited for the task. We hypothesize that this could be  due to the amount or the diversity of the generated morphological inflection training files.

As our systems rely on the output of the previous 3 steps of the baseline, only few training examples were available for Basque and Navajo: 85 and 17, respectively. Probably at least partially due to this fact, i.e., due to finding patterns in the raw text corpus being difficult, all systems obtain their lowest scores on these two languages. However, even though Finnish has 2306 training instances for morphological inflection, our best system surprisingly only reaches $5.38\%$ BMAcc. The same happens in Kannada and Turkish: the inflection training set is relatively large, but the overall performance on unsupervised morphological paradigm completion is low. On the contrary, even though English has a relatively small training set (343 examples), the performance of \IMStwo{} is highest for this language, with $66.20\%$ BMAcc. We think that the quality of the generated inflection training set and the correctness of the predicted paradigm size of the languages are the main reasons behind these performance differences. Improving steps 1 to 3 in the overall pipeline thus seems important in order to achieve better results on the task of unsupervised morphological paradigm completion in the future. 

\section{Conclusion}
In this paper, we described the IMS--CUBoulder submission to the SIGMORPHON 2020 shared task on unsupervised morphological paradigm completion.
We explored two modifications of the official baseline system by substituting its inflection generation component with two alternative models. Thus, our final system performed 4 steps: edit tree retrieval, additional lemma retrieval, paradigm size discovery, and inflection generation. The last component was either an LSTM sequence-to-sequence model with attention (\IMSone{}) or a pointer-generator network (\IMStwo).
Although our systems could not outperform the official baselines on average, \IMStwo{} was the best submitted system. It further obtained the overall highest performance for Bulgarian and Kannada.

\section*{Acknowledgments}
Thanks to Arya McCarthy, Garrett Nicolai, and Mans Hulden for (co-)organizing this shared task, and to Huiming Jin, Liwei Cai, Chen Xia, and Yihui Peng for providing the baseline system!
This project has benefited from financial support to MM by DAAD via a Doctoral Research Grant.

\bibliography{anthology,acl2020}
\bibliographystyle{acl_natbib}

\end{document}